\documentclass[letterpaper, 10 pt, conference]{ieeeconf}
\IEEEoverridecommandlockouts                              
\usepackage{times}
\usepackage{epsfig}
\usepackage{graphicx}
\usepackage{amsmath}
\usepackage{amssymb}

\usepackage{graphicx}
\usepackage{epsfig}
\usepackage{epic,eepic}
\usepackage{times}
\usepackage{mathptmx}
\usepackage{amsfonts}
\usepackage{amsmath}
\usepackage{cite}
\usepackage{color}

\usepackage{times}

\definecolor{red}{rgb}{1,0,0}
\definecolor{green}{rgb}{0,1,0}
\definecolor{blue}{rgb}{0,0,1}
\definecolor{violet}{rgb}{1,0,1}
\definecolor{cyan}{cmyk}{1,0,0,0}
\definecolor{magenta}{cmyk}{0,1,0,0}
\definecolor{yellow}{cmyk}{0,0,1,0}

\definecolor{white}{rgb}{1,1,1}

\usepackage{jumoline}

\newcommand{\CO}[1]{}

\newcommand{\CommentOut}[1]{}

 \newcommand{\editage}[1]{}

\newcommand{\FIG}[3]{
\begin{minipage}[b]{#1cm}
\begin{center}
\includegraphics[width=#1cm]{#2}\\
{\scriptsize #3}
\end{center}
\end{minipage}
}

\newcommand{\FIGR}[3]{
\begin{minipage}[b]{#1cm}
\begin{center}
\includegraphics[angle=-90,clip,width=#1cm]{#2}
\\
{\scriptsize #3}
\vspace*{1mm}
\end{center}
\end{minipage}
}

\newcommand{\acprPaperID}{25}

\begin{document}

\author{\\
\\
\\
{\tt\small ~}
\and
\\
\\
\\
{\tt\small ~}
}

\newcommand{\SW}[2]{#1}

\newcommand{\figA}{
\begin{figure}[t]
  \begin{center}
\FIG{7}{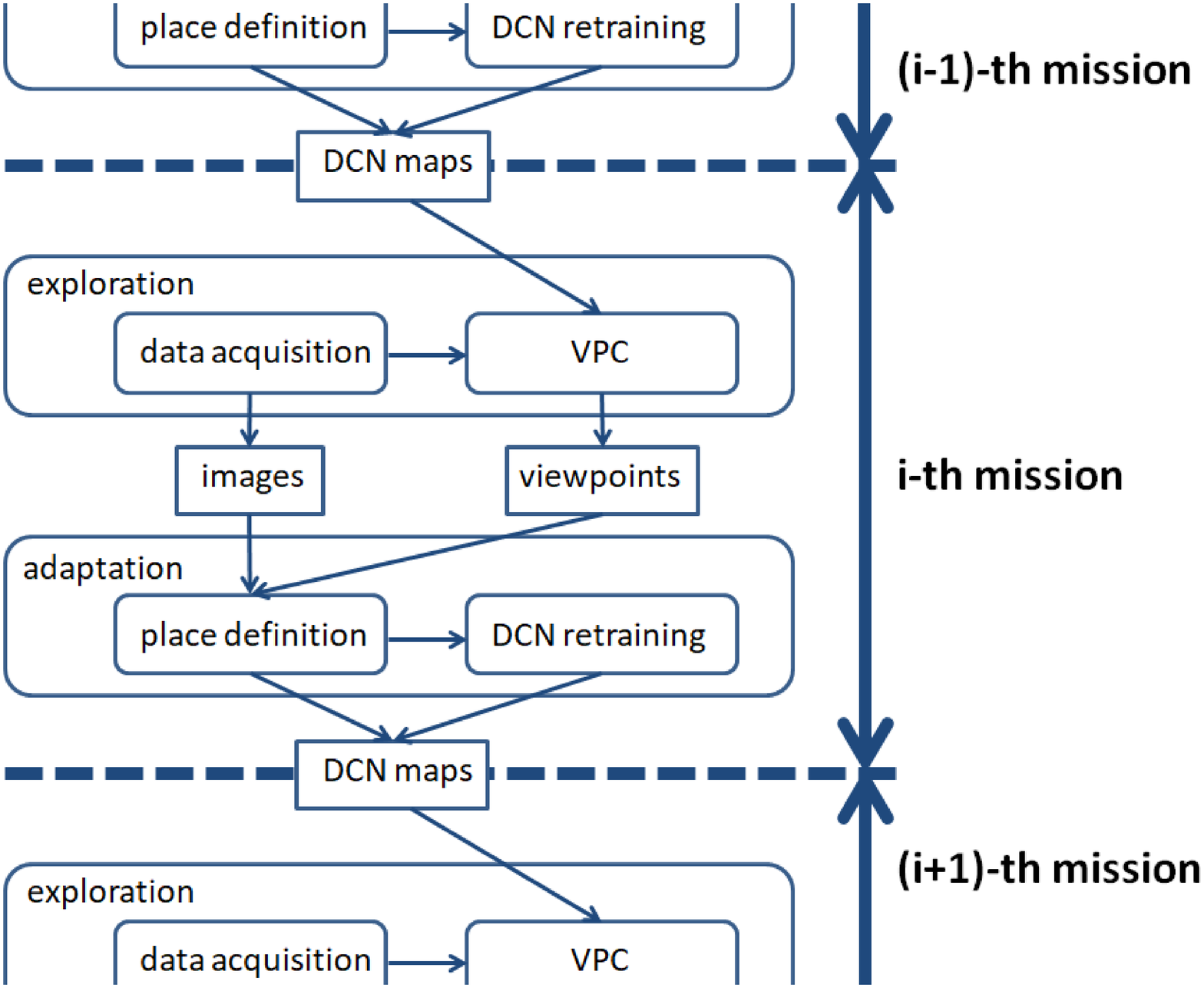}{}
\SW{
\caption{Long-term map learning framework.}
}{

}
\label{fig:A}
\vspace*{-5mm}
\end{center}
\end{figure}
}

\newcommand{\figBa}{
\begin{figure}[t]
  \begin{center}
\FIG{8}{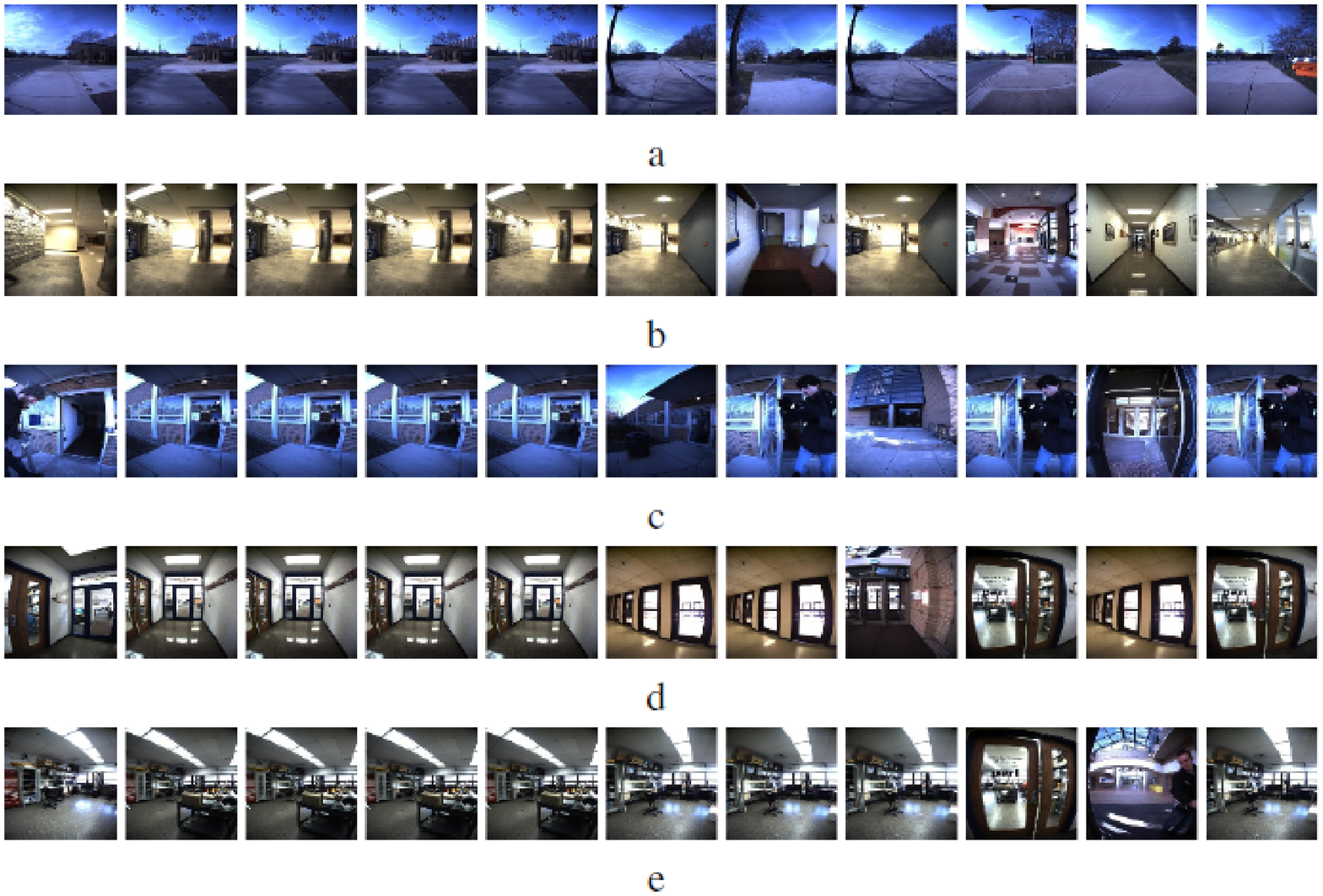}{}
\SW{
\caption{Success examples. For each example, from left to right, the query image, followed by the 1st ranked image through the 10th ranked image. }\label{fig:Ba}
}{

}
\end{center}
\end{figure}
}

\newcommand{\figBb}{
\begin{figure}[t]
  \begin{center}
\FIG{8}{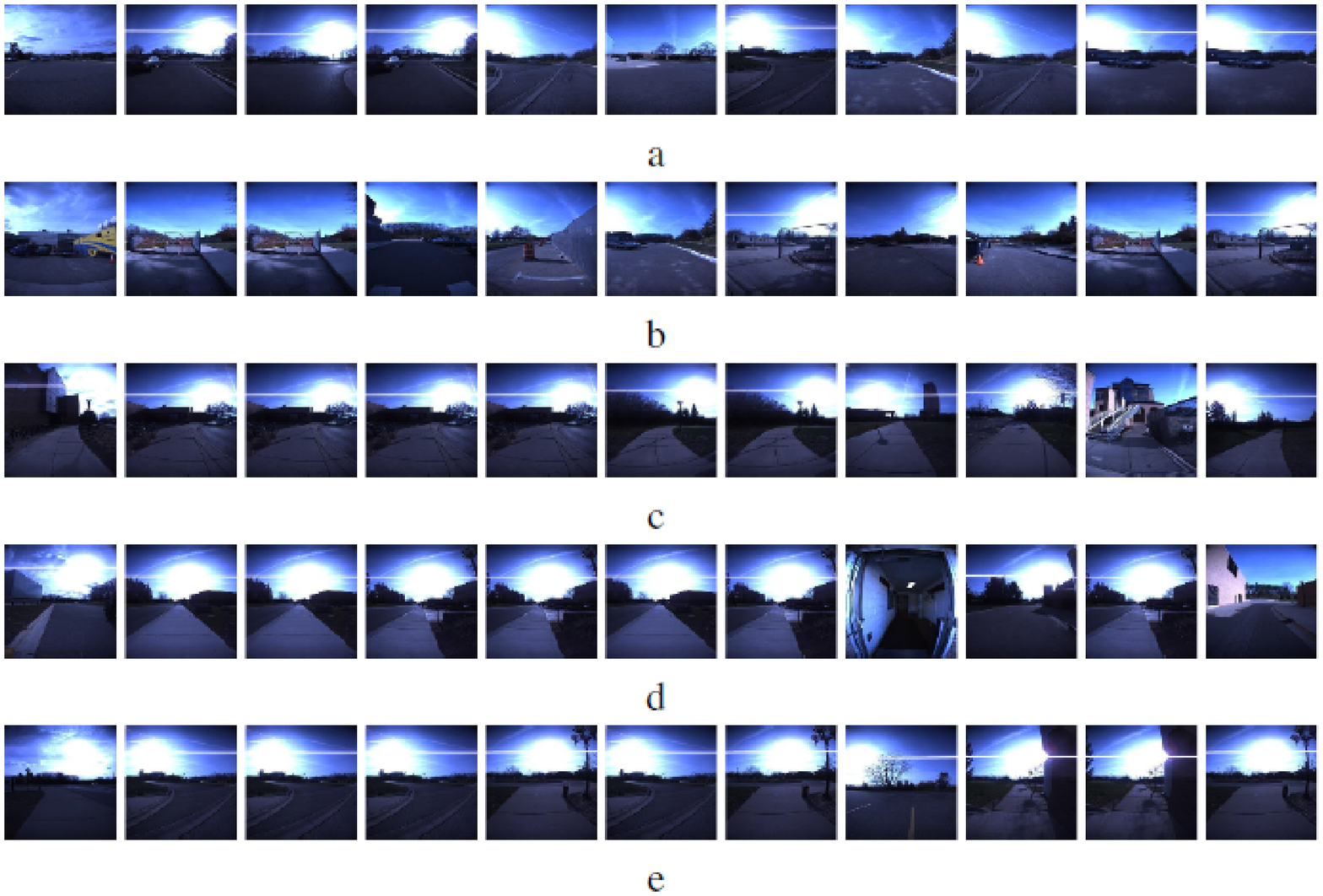}{}
\SW{
\caption{Failure examples. For each example, from left to right, the query image, followed by the 1st ranked image through the 10th ranked image.}\label{fig:Bb}
}{

}
\end{center}
\end{figure}
}

\newcommand{\figC}{
\begin{figure}[t]
  \begin{center}
\FIG{8}{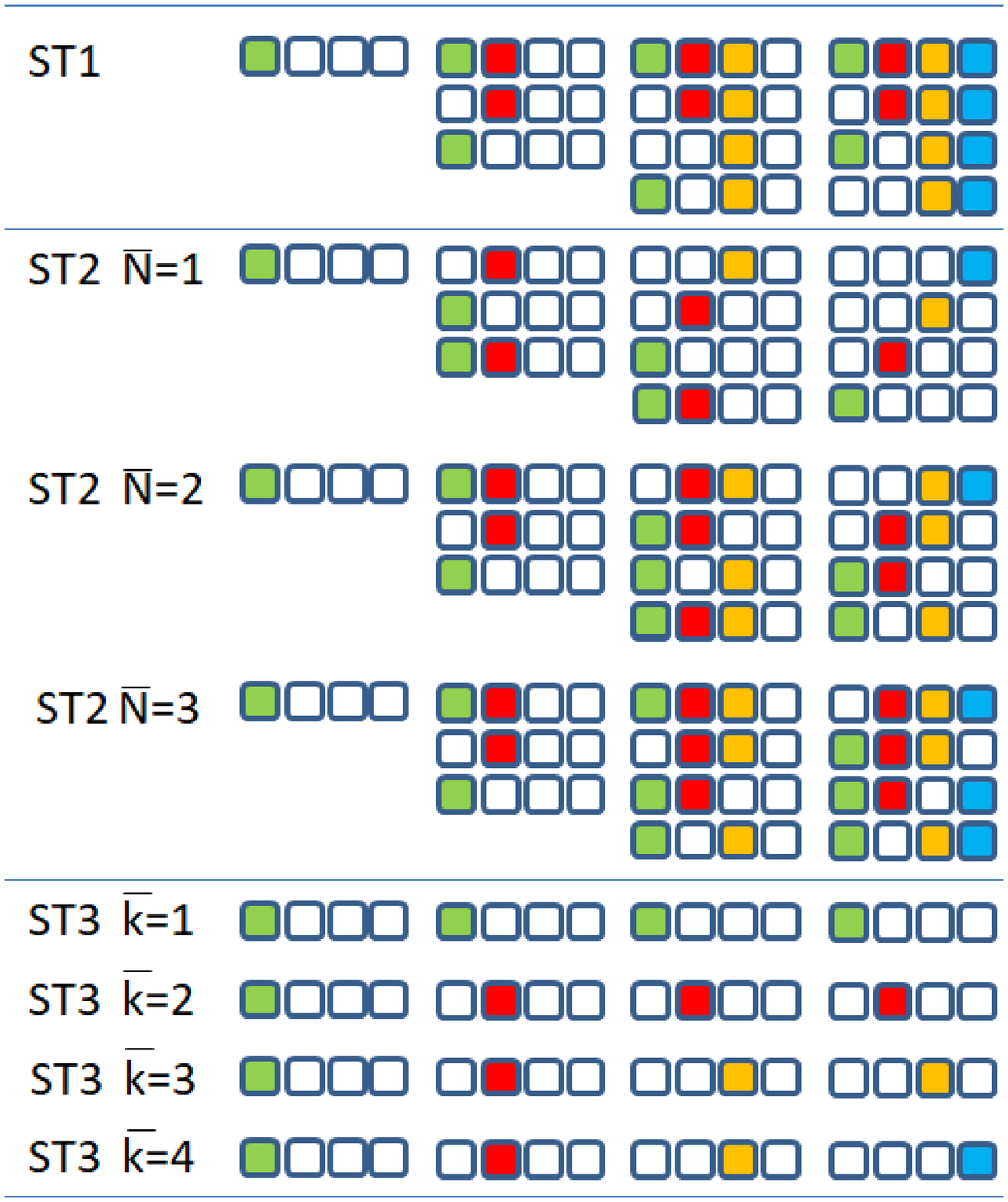}{}
\SW{
\caption{Visualization of different scheduling strategies.
For each method, 
each row corresponds to a specific DCN.
Each column corresponds to a specific training set,
from left to right,
``3/31," ``8/4," ``11/17," and ``1/22"
respectively.
Each colored box 
(green, red, orange, blue)
indicates 
that a specific DCN is fine-tuned by a specific training set
(``3/31," ``8/4," ``11/17," ``1/22").
}\label{fig:C}
}{

}
\end{center}
\end{figure}
}

\newcommand{\figE}{
\begin{figure}[t]
  \begin{center}
\FIGR{6}{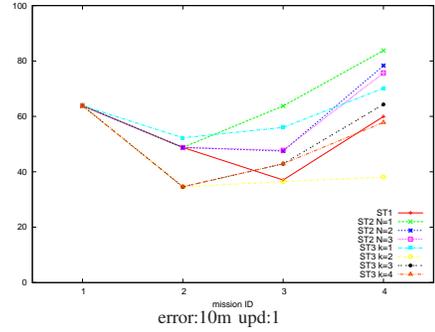}{error:10m upd:1}
\FIGR{6}{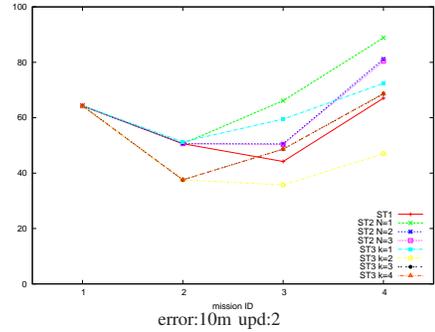}{error:10m upd:2}
\FIGR{6}{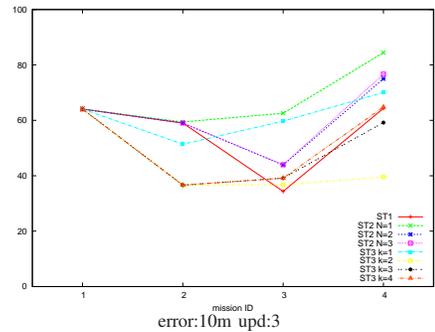}{error:10m upd:3}\\
\SW{
\caption{Performance results. Vertical axis: success ratio. Horizontal axis: mission ID. The mission IDs $j$ = 1, 2, 3, and 4 respectively correspond to the seasons of ``3/31," ``8/4," ``11/17," and ``1/22." ``error" indicates the location error allowed [m] that is used 
to judge whether a VPC task is successful or not to compute success ratio.
``upd:1, 2, and 3" respectively indicate the place definition strategies ``location," ``location-appearance," and ``incremental clustering" described in \ref{sec:upd}.}\label{fig:E}
}{
}
\end{center}
\end{figure}
}

\newcommand{\figF}{
\begin{figure}[t]
  \begin{center}
\FIGR{2.7}{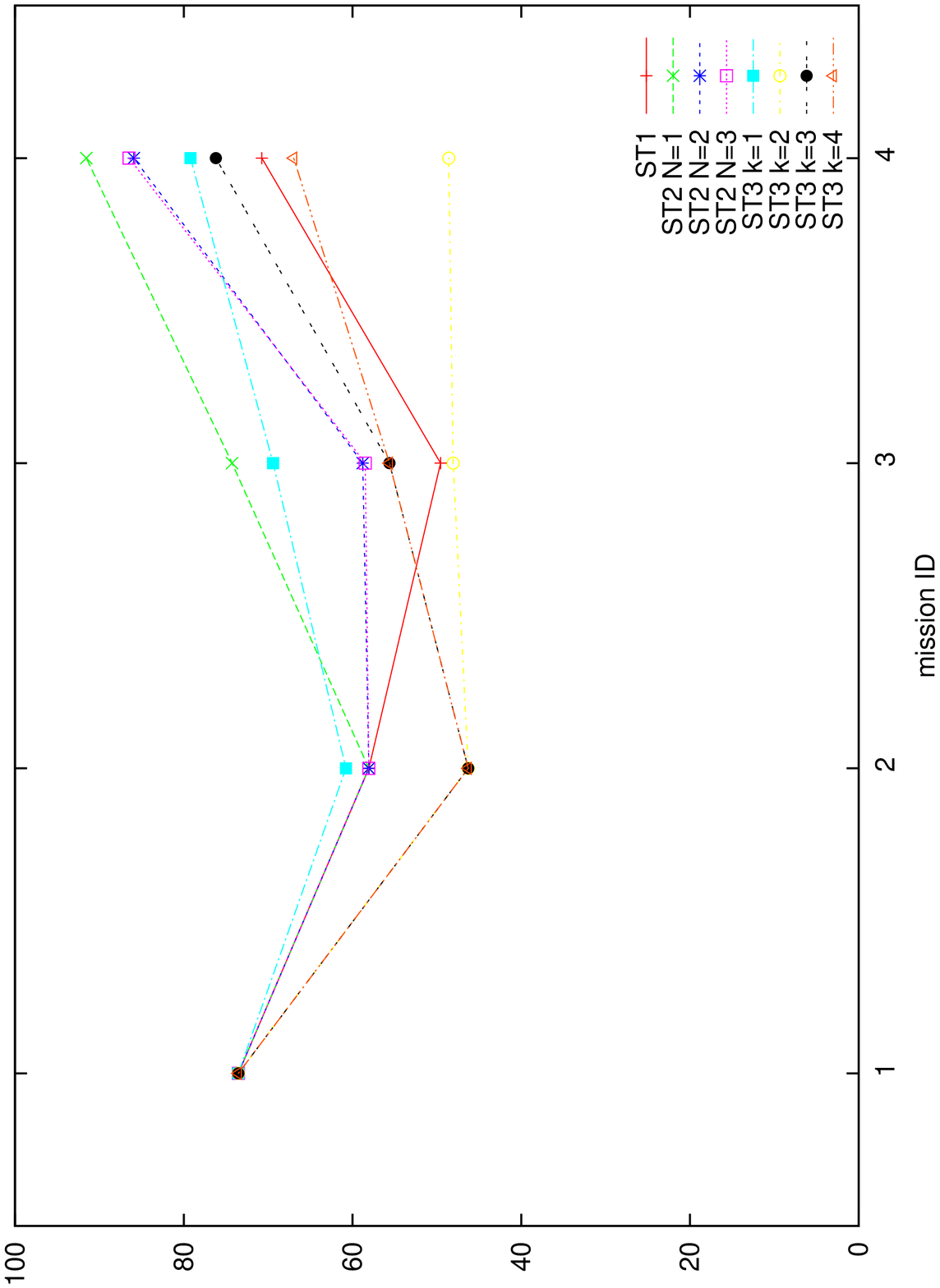}{err:20m upd:1}\hspace*{-3mm}%
\FIGR{2.7}{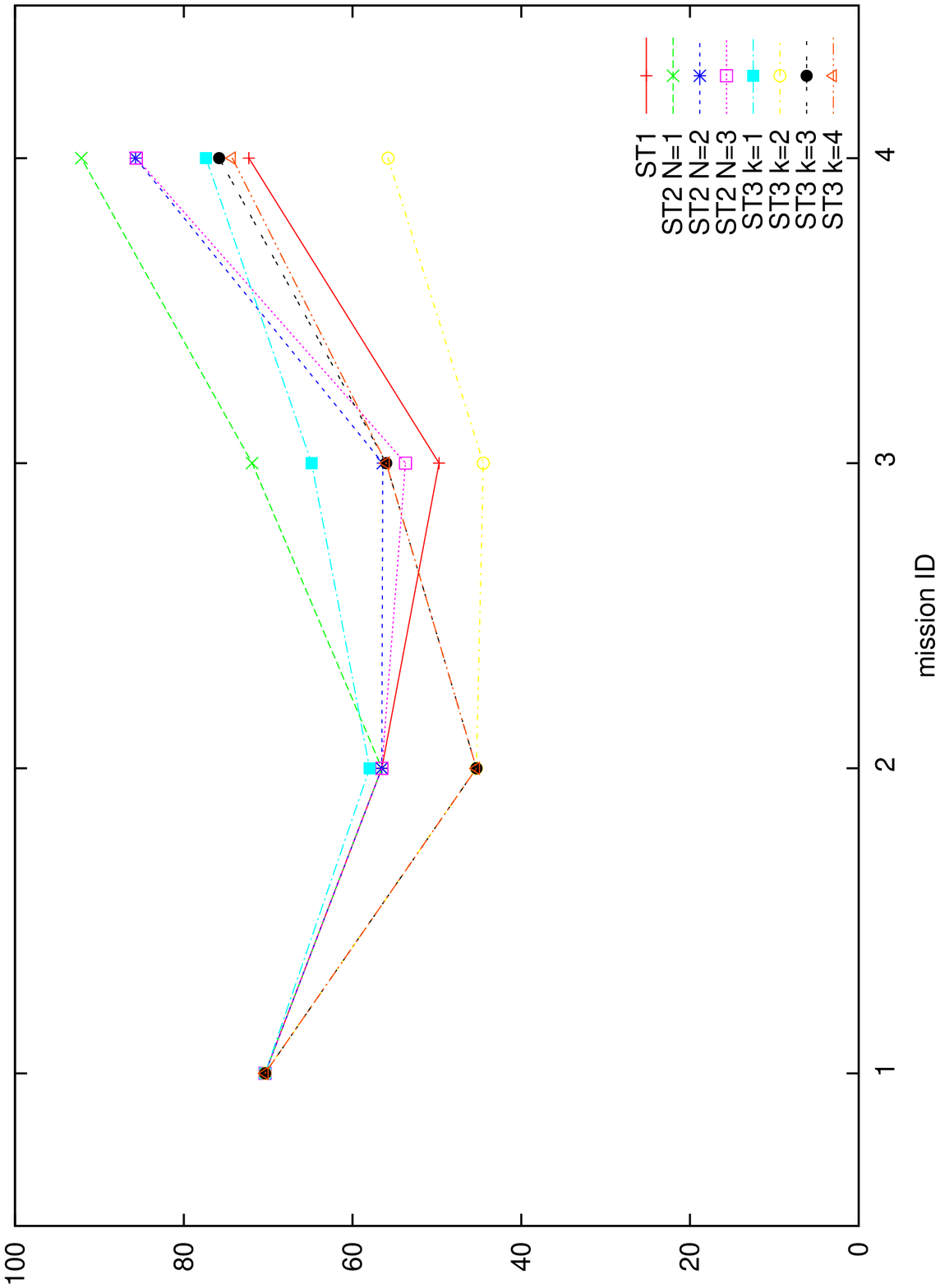}{err:20m upd:2}\hspace*{-3mm}%
\FIGR{2.7}{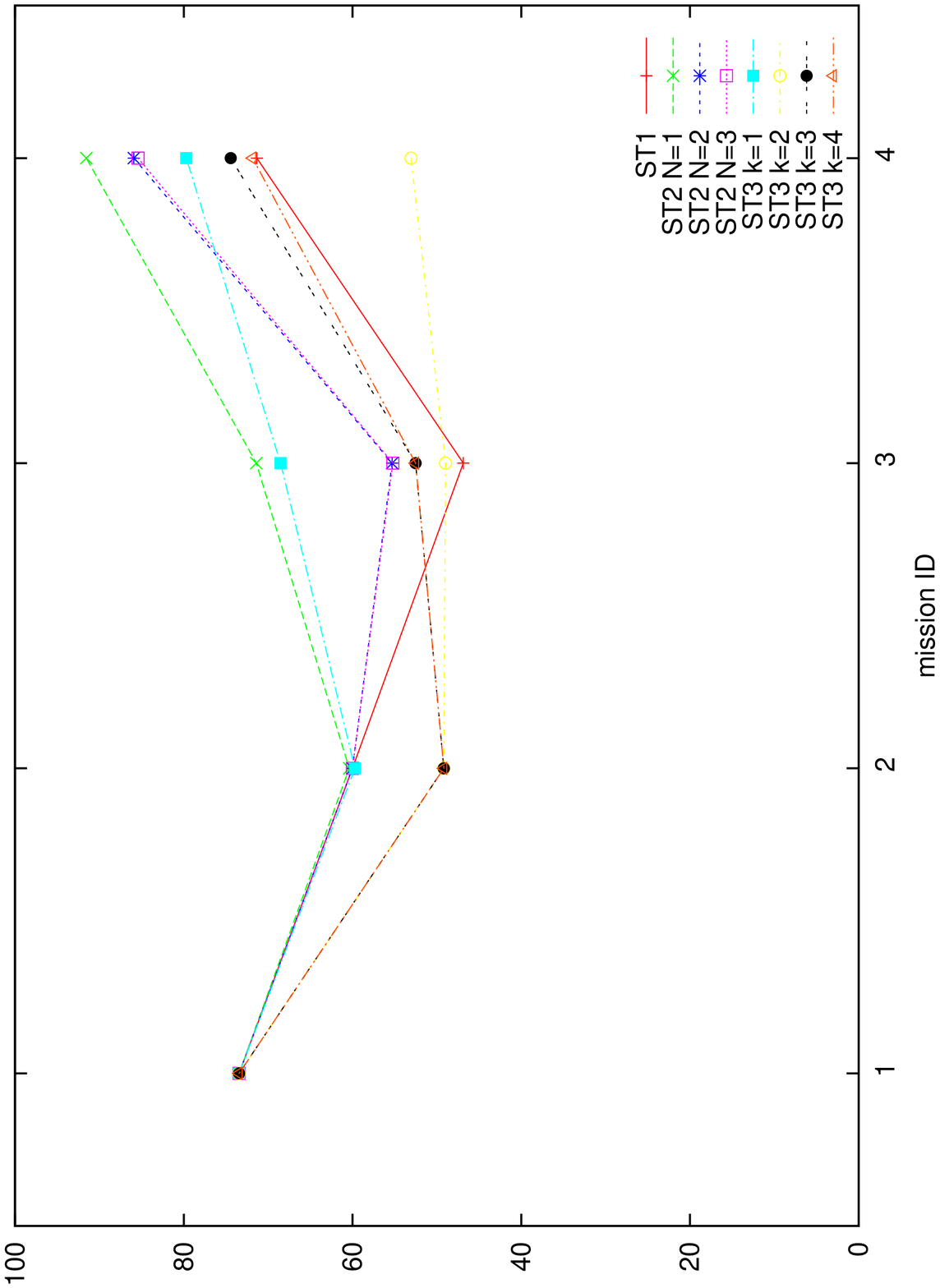}{err:20m upd:3}\hspace*{-3mm}\\
\FIGR{2.7}{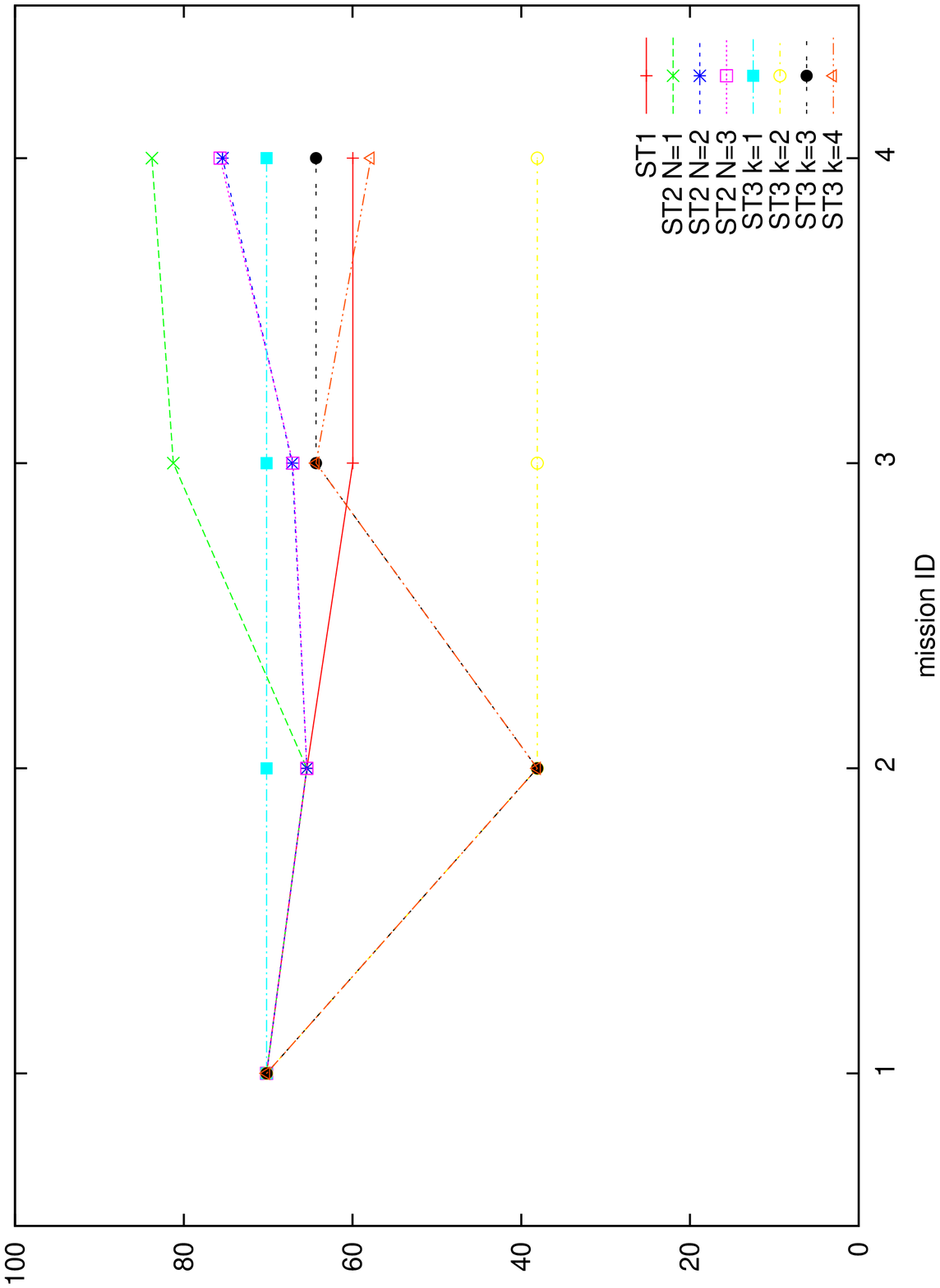}{err:10m test:ex upd:1}\hspace*{-3mm}%
\FIGR{2.7}{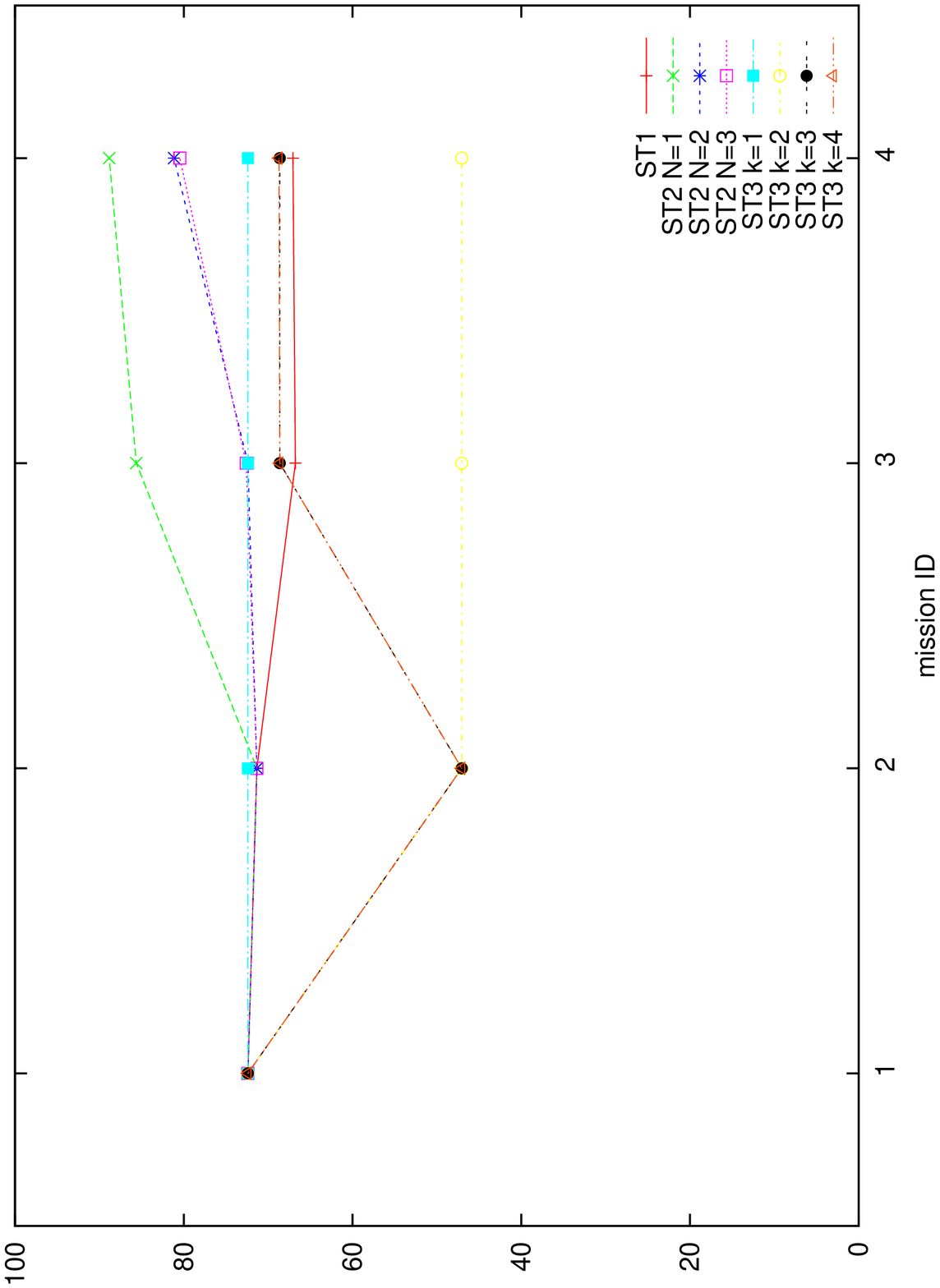}{err:10m test:ex upd:2}\hspace*{-3mm}%
\FIGR{2.7}{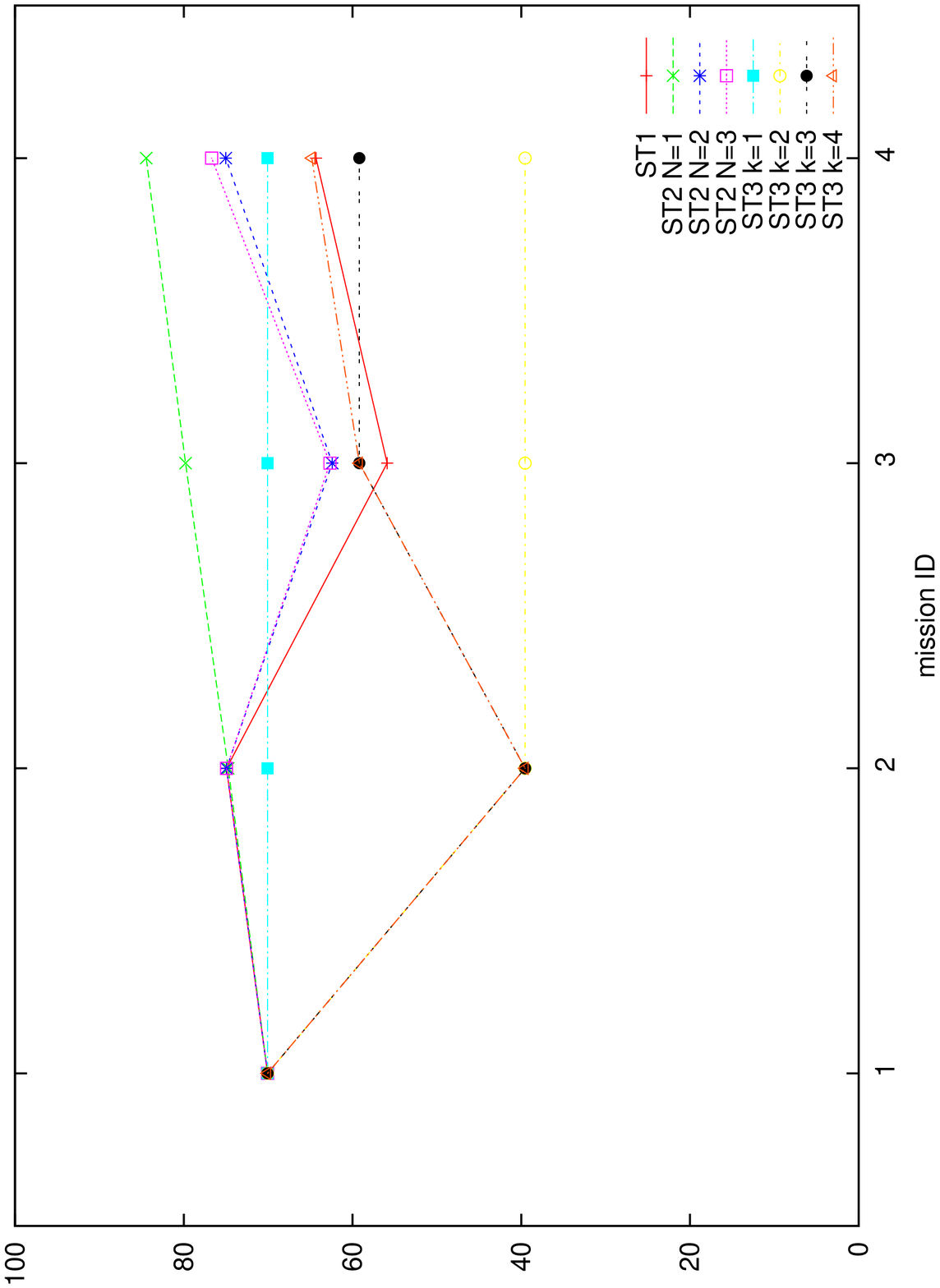}{err:10m test:ex upd:3}\hspace*{-3mm}\\
\FIGR{2.7}{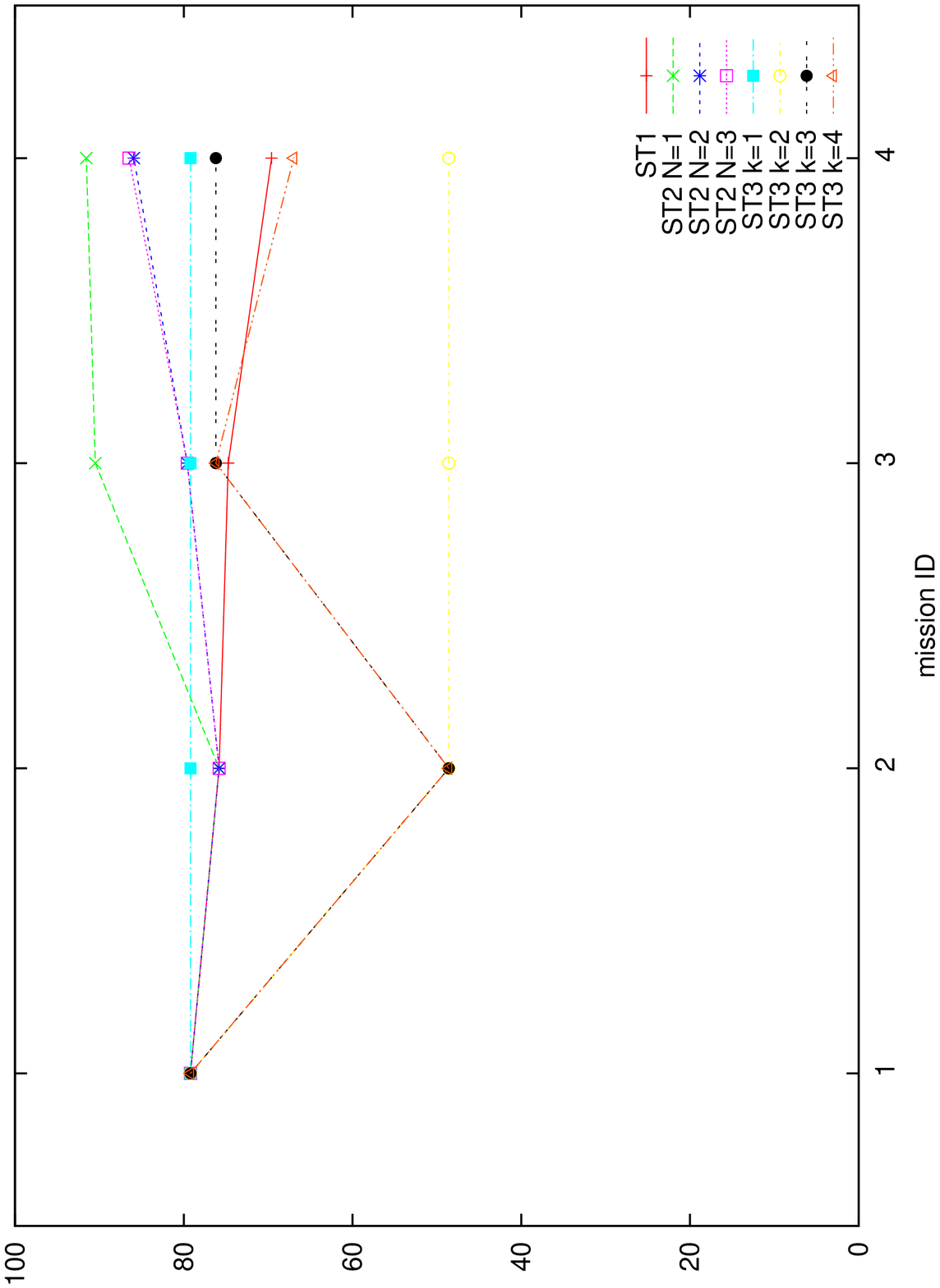}{err:20m test:ex upd:1}\hspace*{-3mm}%
\FIGR{2.7}{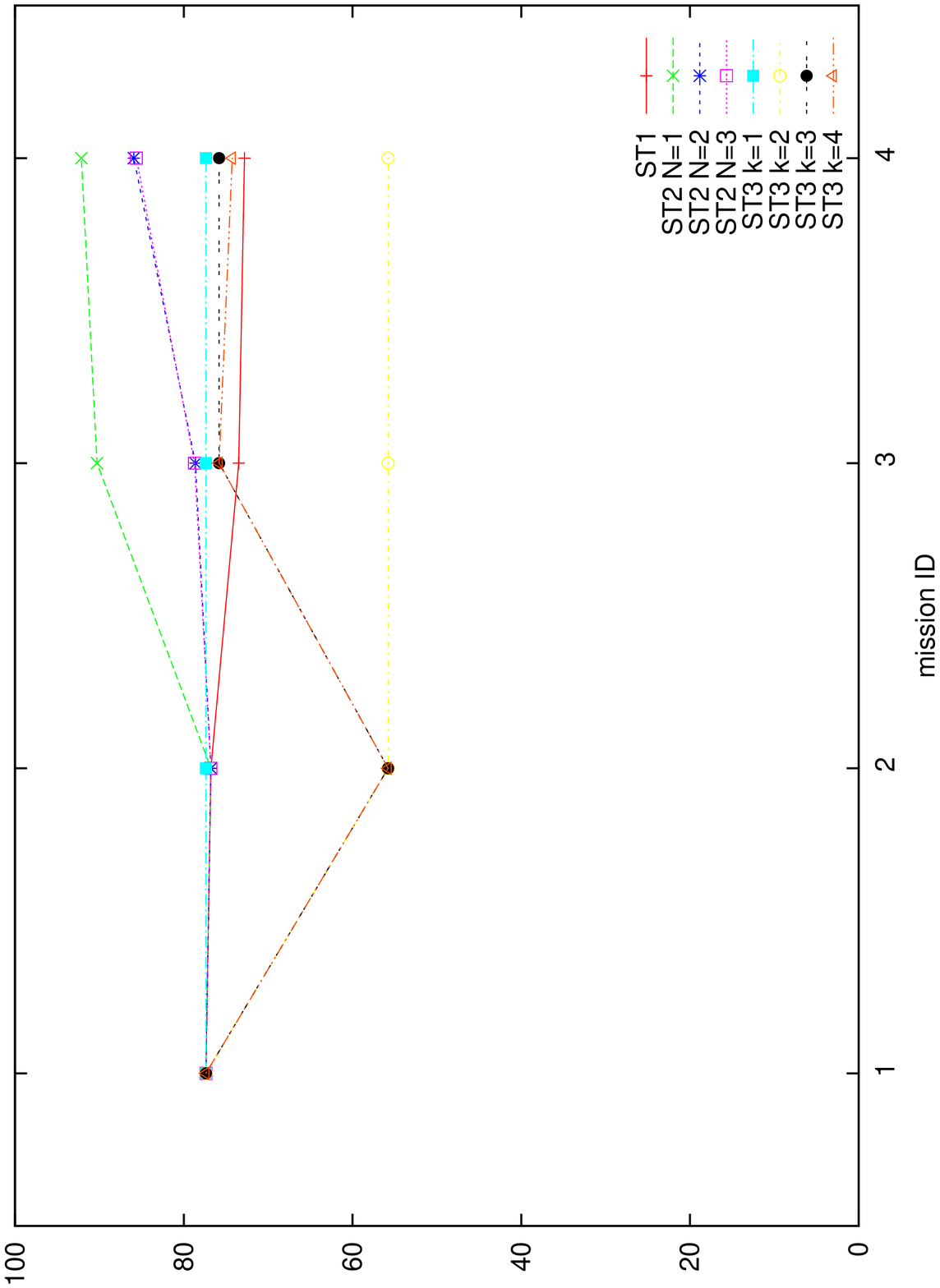}{err:20m test:ex upd:2}\hspace*{-3mm}%
\FIGR{2.7}{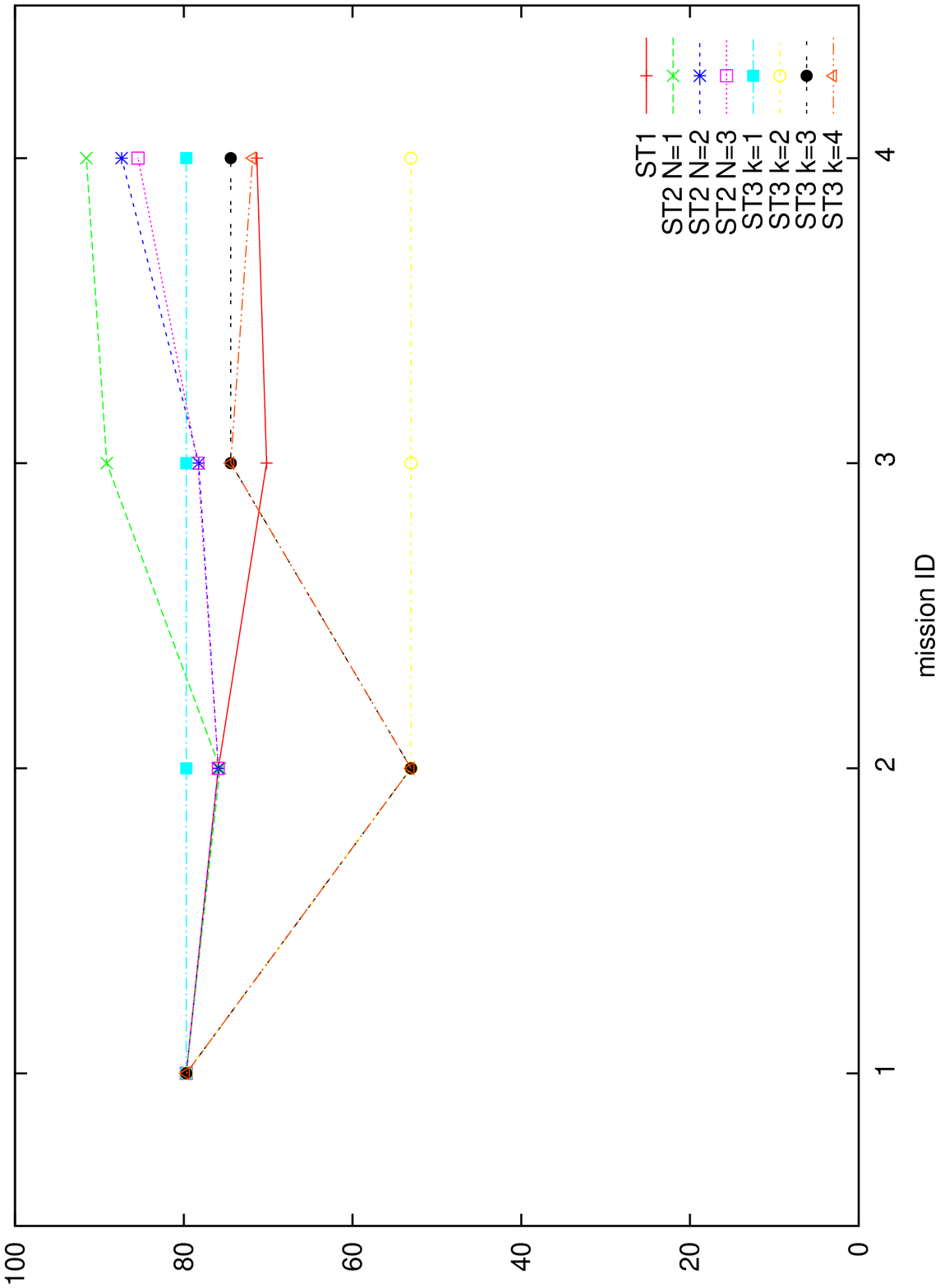}{err:20m test:ex upd:3}\hspace*{-3mm}\\
\SW{
\caption{Additional results using different settings.}\label{fig:F}
}{
}
\end{center}
\end{figure}
}

\newcommand{\figG}{
\begin{figure}[t]
  \begin{center}
\FIG{8}{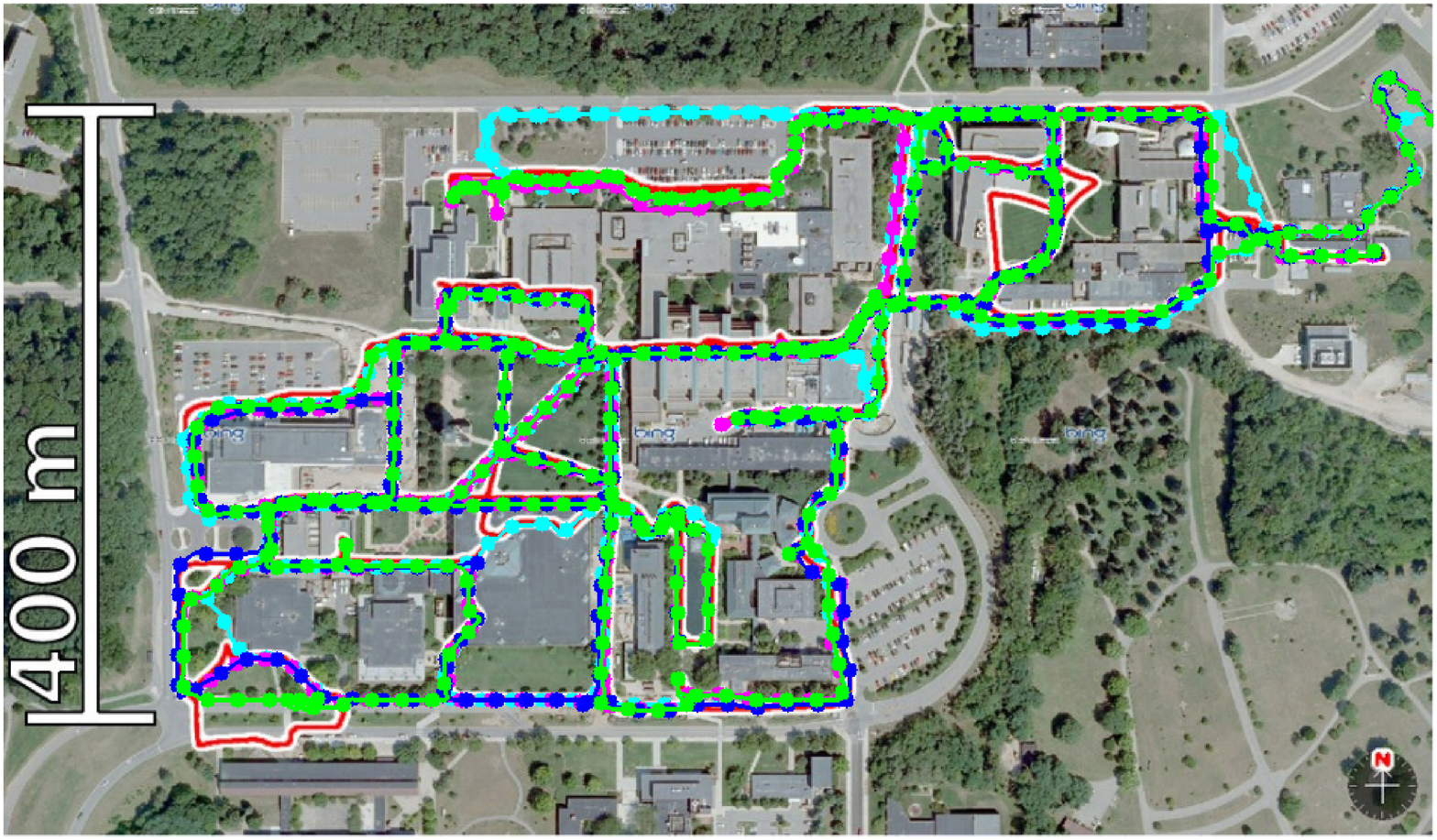}{}
\SW{
\caption{Experimental environment. The trajectories of the four training sets, ``2012/3/31," ``2012/8/04," ``2012/11/17," and ``2012/1/22," 
used in our experiments are visualized in 
purple, blue, light-blue, and green 
curves and are overlaid using a bird's eye view imagery.}\label{fig:G}
}{

}
\end{center}
\end{figure}
}

\newcommand{\figH}{
\begin{figure*}[t]
  \begin{center}
\FIG{17}{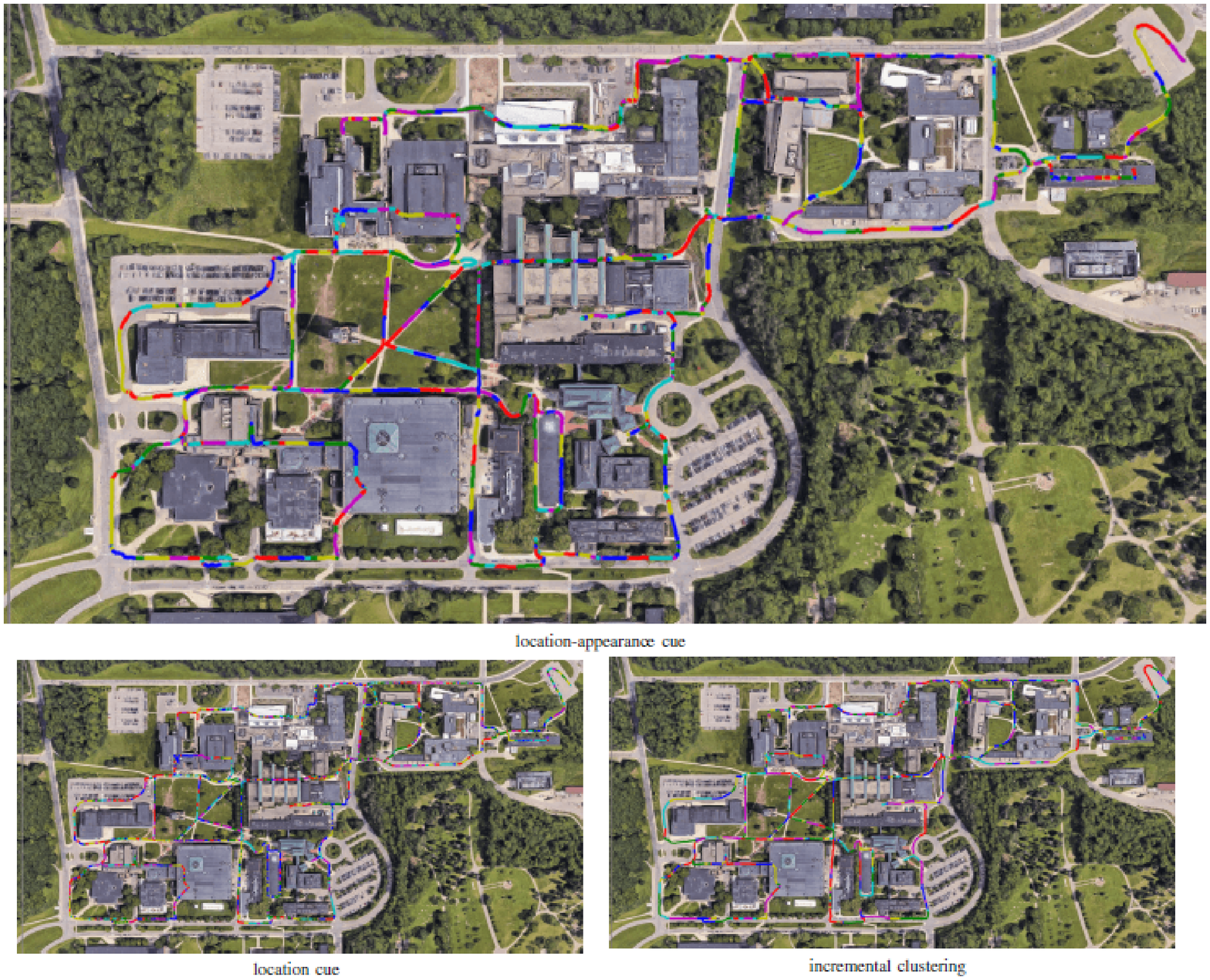}{}
\SW{
\caption{Qualitative results of unsupervised place definition algorithms. Locations on the robot's trajectories are 
classified into different place classes
and overlaid on bird's eye view imagery of the environment
using different colors for different classes.}\label{fig:H}
}{

}
\end{center}
\end{figure*}
}

\newcommand{\figI}{
\begin{figure}[t]
  \begin{center}
\FIG{8}{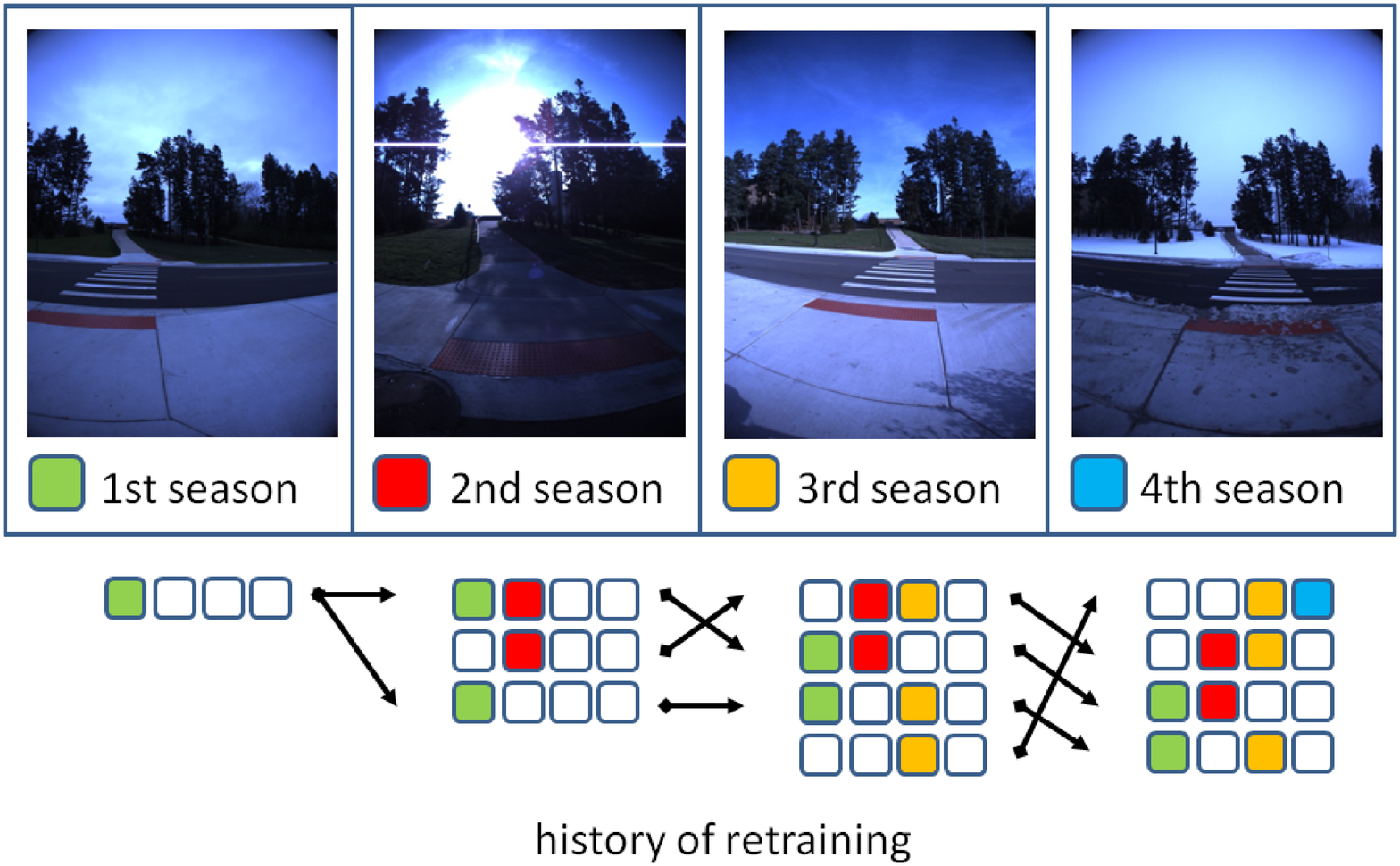}{}\vspace*{-2mm}\\
\SW{
\caption{Long-term map learning for cross-season visual place classification. To realize a good tradeoff between generalization and specialization abilities, we employ an ensemble of DCN classifiers and consider the task of scheduling when and which classifiers to retrain, given previous season's DCN classifiers as the sole prior knowledge. }\label{fig:I}
}{

}
\end{center}
\vspace*{-8mm}
\end{figure}
}

\title{\LARGE \bf
Long-Term Ensemble Learning of Visual Place Classifiers
}

\author{~~~~~~Fei Xiaoxiao~~~~~~~~~Tanaka Kanji~~~~~~Fang Yichu~~~~~~Takayama Akitaka
\thanks{Our work has been supported in part by 
JSPS KAKENHI 
Grant-in-Aid 
for Scientific Research (C) 26330297, and (C) 17K00361.}
\thanks{The authors are with Graduate School of Engineering, University of Fukui, Japan. 
{\tt\small tnkknj@u-fukui.ac.jp}}}

\maketitle

\begin{abstract}
\SW{
This paper addresses the problem of cross-season visual place classification (VPC) from a novel perspective of long-term map learning. Our goal is to enable transfer learning efficiently from one season to the next, at a small constant cost, and without wasting the robot's available long-term-memory by memorizing very large amounts of training data. To realize a good tradeoff between generalization and specialization abilities, we employ an ensemble of convolutional neural network (DCN) classifiers and consider the task of scheduling (when and which classifiers to retrain), given a previous season's DCN classifiers as the sole prior knowledge. We present a unified framework for retraining scheduling and discuss practical implementation strategies. Furthermore, we address the task of partitioning a robot's workspace into places to define place classes in an unsupervised manner, rather than using uniform partitioning, so as to maximize VPC performance. Experiments using the publicly available NCLT dataset revealed that retraining scheduling of a DCN classifier ensemble is crucial and performance is significantly increased by using planned scheduling.
}{
}
\end{abstract}

\section{Introduction}

\SW{
This paper addresses the problem of visual robot self-localization
from a novel perspective of long-term map learning. We follow the recent self-localization paradigm based on a deep convolutional neural network (DCN) \cite{1}. 
Thus, an environment map is learned as a DCN-based visual place classifier, and which is used to classify a query image into one of the learned place classes. We address the difficult long-term scenario of visual place classification (VPC), termed cross-season VPC \cite{2}, 
where training and test images involve different seasons.
One of most basic schemes to handle this difficulty, is to train a DCN classifier from all available training images. 
However, this requires a robot to explicitly memorize and learn a number of training images proportional to the number of places and seasons, which severely limits the scalability of the algorithm in both time and memory space. 
}{
}

\SW{
Our goal is to develop a long-term map learning framework that enables efficient retraining of the VPC system, at a small constant cost, and without wasting the robot's available long-term-memory by memorizing very large amounts of training data.
This study is inspired by recent progress in domain adaptation and transfer learning \cite{20151009-1,3,7,8,9,20151030-1,20151204-1}, 
where the aim is to learn a classifier model for a target domain by exploiting rich information present in a source domain. In our study, classifiers learned in previous seasons represent the source knowledge, and we aim to exploit the source knowledge to improve the current season's VPC performance.
We follow the literature of domain adaptation and transfer learning,
although a key difference is that
in our application scenario of autonomous robotics,
the definitions of place classes and domains 
are not provided,
and the robot must discover their optimal definitions in an unsupervised manner. 
}{

}

\figI

\SW{
To obtain an acceptable tradeoff between generalization and specialization, it is crucial to adequately train and retrain DCN classifiers (Fig. \ref{fig:I}). Thus, if a DCN classifier is retrained (i.e., fine-tuned) to a specific season's training data, its specialization ability is expected to increase, but its generalization ability tends to decrease. Thus, we have two possible choices: to either retrain a specific DCN classifier with a specific training set or not. After collecting $N$ different training sets, there are $2^N$ possible choices and $2^N$ possible DCN classifiers. Training and using this exponential number of DCN classifiers is often intractable. We suggest a solution based on ensemble learning that requires only a fixed set of classifiers that integrate information from multiple domains using fine-tuning and classifier fusion. 
}{
}

\SW{
More formally, we address two different questions. 
The first question is how to choose which DCN classifiers to retrain, with the current season's training set, out of the available DCN classifiers trained in the previous season. Recent advances in fine-tuning techniques for DCN have simplified the retraining task \cite{20151009-1}. 
However, there is no straightforward method for retraining scheduling that achieves an optimal tradeoff between VPC accuracy and training efficiency. 
Secondly, we address the question of how to integrate the outputs from multiple classifiers from different seasons. Because individual classifiers are trained using different amounts of training data from different seasons, they often provide conflicting classification results with different levels of variances. The key to this question is to fuse probability estimates from individual DCN classifiers. In this study, we present and evaluate several strategies for retraining scheduling and for applying classifier fusion.
}{
}

\SW{
An open question is how to partition the robot's workspace into places to define place classes. 
Intuitively, each place class should be defined as a continuous region in the robot's workspace with similar visual features. 
The main difficulty is the chicken and egg problem: If we have a well-trained classifier, it is rather easy to partition the robot's workspace into place regions, but the training of a classifier requires a set of pre-defined place classes. 
Optimal definition of places is a practical concern, as the definition of place classes strongly influences VPC performance. It simplifies the inference over the space of robot pose and enables efficient self-localization, by using non-uniform planned partitioning of the space, as opposed to typical uniform partitioning. From a broader perspective, optimal place definition is interesting because it may facilitate a unifying framework for compact map representation. 
In this study, we present several strategies 
for the place-definition and workspace-partitioning discovery.
}{
}

\SW{
The main contribution of this study is an extension of the VPC framework to setups with long-term map learning. 
This study is inspired by our previous studies on 
cross-season self-localization \cite{icra15kanji}
and DCN-based localization \cite{iros16kanji};
however,
a key novelty is the formulation of cross-season VPC.
The optimal definition of place classes is inspired by our previous study in \cite{mva17kanji};
however, the additional problem of domain adaptation between seasons arises in the long-term map learning scenario. We address this important issue and present a practical solution for it. We follow the literature such as \cite{20151009-1} that suggests the use of the Alexnet architecture to analyze transfer learning and we focus on the problem of when and which DCN classifiers to retrain so as to maximize performance of the ensemble classifier.
Our experimental results using the publicly available NCLT dataset \cite{6} revealed that retraining scheduling of a DCN classifier ensemble is crucial and performance is significantly increased using planned scheduling.
}{
}

\section{Related Works}

\SW{
End-to-end training of DCN for visual self-localization has attracted interest in recent years. 
In \cite{posenet}, DCN is introduced as the regressor to achieve end-to-end training of 6DOF camera relocalization using RGB and RGB-D. Very recently \cite{12}, 
the two regression approaches of random forest (RF) \cite{RF4VPR} 
and DCN are compared,
and furthermore the novel task of mapping RF to a neural network is considered to achieve a good efficiency-accuracy tradeoff. 
On the other hand, a major limitation of regression approaches is that they require fine-grained training sets, such as images annotated with 6DOF camera poses, which severely limits their applicability to large-scale long-term map learning. 
One of the most similar formulations to ours is the formulation of topological localization \cite{13}, which has some desirable properties including map compactness and robustness against map errors.
}{
}

\SW{
Alexnet has been a popular tool for analyzing transfer learning. In \cite{20151009-1}, 
analysis of transfer learning, rather than achieving state-of-the-art performance, is the main focus.
The reference implementation by Caffe is used in its original form, so that the analysis results will be comparable, extensible and useful to larger number of researchers.%
}{

}
\SW{
In \cite{20170627-1}, 
the authors argue that a large image dataset such as ImageNet contains much more information than officially announced, and most often such existing knowledge resources are ignored. 
Based on this idea, they presented a novel method for  zero-shot learning (i.e., transfer learning).
In \cite{20170124-1}, 
the problem of topological self-localization 
is addressed using fusion and binarization of DCN features.
In the study, the DCN architecture is based on a pre-trained model using the ImageNet dataset, to confirm the generalization of the automatically learned features, and to demonstrate that the description power acquired by the DCN is transferable to specific datasets.
}{
}

\SW{
Our approach is informed by domain adaptation and transfer learning approaches, ranging from parameter adaptation, feature transformation, and metric learning, to deep learning techniques, 
which have been applied to wide variety of visual recognition tasks \cite{8}. 
}{

}
\SW{
In \cite{4},
a feature transformation termed marginalized denoising autoencoder (MDA)
has been extended to denoise both the source and target data in such a way that the features become domain invariant and adaptation is easier.
In \cite{9},
scalable greedy algorithms for transfer learning 
are presented, where the authors focus on how to select and combine sources from a large pool of data to yield good performance on a target task.
In \cite{20151030-1}, 
the problem of classifier learning from only positive and unlabeled data is addressed on binary classifier (e.g., SVM), and exploit the fact that the conditional probability of a model trained on labeled and unlabeled examples is not very different from a model trained on fully labeled examples, assuming that positive examples are labeled at random.
}{
}
\SW{
In \cite{20170622-1}, 
the problem of transfer learning is addressed
in an interesting setting, where the target class has very few training examples. The authors aim to discover similar classes and transfer knowledge among them, by assuming that the classes have been organized into a fixed tree hierarchy and that the hierarchy is available or learnable. 
}{
}

\SW{
Our study is related to the paradigm of life-long learning or open world recognition,
in which knowledge is accumulated and maintained across domains.
}{
}
\SW{
We also employ mid-level image representation provided by DCN.
In \cite{14}, 
the authors present a novel region-based image representation where the Naive Bayes nearest neighbor model is applied and seamlessly integrated into a DCN.
Very recently \cite{chen2017only},
a new region-based feature encoding 
is presented using multiple convolutional layers for feature extraction and saliency identification.
}{

}
\SW{
Our approach is also related to ensemble learning of DCNs; 
However, use of DCN ensembles in visual self-localization has not been explored in the context of long-term map learning.
In this study, we present a novel DCN ensemble approach that is specifically customized for visual place classifiers.
}{

}

\section{Approach}

\SW{
The long-term map learning framework consists of two alternately repeated missions (one iteration): exploration and adaptation (Fig. \ref{fig:A}). 
The framework is initialized with a size one classifier set $C^0$$=\{C_1^0\}$,
which consists of a single DCN classifier $C^0_1$ that is obtained by pretraining a DCN using Bigdata such as ImageNet. A new classifier set $C^i=$$\{C_j^i\}$ is then obtained by using additional training data in each $i$-th iteration ($i\ge 1$). 
In experiments, we use as the initial DCN classifier $C_1^0$ the Alexnet architecture pretrained on the ImageNet LSVRC-2012 dataset, and we consider one iteration of the two missions per season. 
}{
}

\SW{
The exploration mission aims at robot exploration of the entire environment, while keeping track of the robot's global position (e.g., using pose tracking and relocation), as much as possible, in order to collect mapped images that have global viewpoint information, and optionally, the collected data may be further post-processed to refine the viewpoint information by structure-from-motion \cite{15} or SLAM \cite{16}. 
All the collected images that have viewpoint information are used as training data for the subsequent $i$-th adaptation mission (See Fig. \ref{fig:A}). 
We denote training data that is collected in the $i$-th exploration as $D^i=\{(v, I)\}$, where $I$ and $v$ respectively are an image and its viewpoint.
}{
}

\SW{
The adaptation mission aims to obtain a new set of DCN classifiers $C^i$ by fine-tuning existing DCN classifiers $C^{i-1}$ based on transfer learning and domain adaptation, given training data $D^i$ that is obtained in the latest $i$-th exploration mission. 
As mentioned previously, we have a binary choice: whether a specific DCN classifier in $C^{i-1}$ should be fine-tuned with a specific training set or not, where there are $2^i$ possible DCN classifiers.
We denote a new classifier that is obtained by fine-tuning an existing classifier $C^{i-1}_j$
by incorporating a new training set $D^i$ as $C^{i-1}_j \oplus D^i$. For example, if we fine-tune a DCN $C_1^0$ using training data $D^3$ and then the resulting DCN is further fine-tuned using $D^4$, the final DCN is $C=C_1^0 \oplus D^3 \oplus D^4$.
We discuss the topic of retraining scheduling (i.e., the questions of when and which DCN classifiers should be fine tuned) in \ref{sec:schedule}.
}{
}

\SW{
The adaptation mission also involves the discovery of a new set of place classes that is suitable for VPC. 
Since the area covered by the robot exploration and its appearance differs among different explorations, the way of defining place classes should also differ among different environments. We discuss the topic of unsupervised place-definition and workspace-partitioning discovery in \ref{sec:upd}.
}{
}

\figA

\SW{
The VPC task is a part of the exploration mission and attempts visual robot localization using the latest classifiers $C^{i-1}$. The VPC task assumes no prior knowledge of the robot pose, which is a challenging self-localization scenario called global localization \cite{13}, although our VPC would also be useful for other scenarios, including pose tracking. Ideally one would like to use only a single classifier $C=C_1^0\oplus_{j=1}^i D^j$ that has been repeatedly fine-tuned using all available training data as it is expected to be most informative among all possible DCN classifiers. However, in practice, this simple strategy turns out to yield poor VPC performance, due to overfitting and numerous false positives. Therefore we apply fusing information $C^*=F(C^i)$ from an ensemble of DCN classifiers to obtain more reliable classification results. The definition of place classes can be different among different classifiers, so transform outputs from individual classifiers to a unified global map coordinate system using a fusion function. We discuss the information fusion function $F$ in \ref{sec:fusion}.
}{
}

\subsection{Retraining Scheduling}\label{sec:schedule}

\SW{
Recall that the $i$-th adaptation mission 
selects a subset of existing DCN classifiers $\{C_1^0\}$ $\cup$ $C^{i-1}$,
retrains (i.e., fine-tunes) each of the selected classifiers using the newly obtained training data $D^i$,
and then replaces one of the existing classifiers with each newly trained one.
Therefore,
we need to schedule which classifier to retrain and which classifier to replace,
given the classifier set $\{C_1^0\}$ $\cup$ $C^{i-1}$.
Note that 
a DCN classifier $k\in [1, |C^i|]$ at the $i$-th mission
can be uniquely identified by its history of retraining
in the $j$-th mission ($j\in [1, i]$).
For simplicity,
let us denote this history by a bit string
$B^i=[b^1 \cdots b^i]$
where each bit $b^j$ $(j\in [1, i])$ 
represents whether the specific DCN classifier
has been retrained ($b^j=1$) or not ($b^j=0$) at the $j$-th adaptation mission with the $j$-th training data.
}{
}

\SW{
In this study, we developed three different strategies for scheduling. 
}{

}

\SW{
The first strategy, termed ST1, is based on the idea that the newest training set (acquired at the current $i$-th season) is expected to be best suited for future missions and hence is preferentially selected for the current mission's retraining. This strategy is represented by
\begin{equation}
\hat{B}^i = arg \max_{B^i} \left( \sum_{k=1}^i B_k^i \cdot [0 \cdots 0 1]^T + \frac{N(B_k^i)}{1+i} 
\right).
\end{equation}
The function $N(B)$ returns the number of 1-bits in $B$
\begin{equation}
N(B) = B \cdot [1 \cdots 11]^T
\end{equation}
and is used here as a lower priority objective for maximizing the number of 1-bits in $B_k^i$.
}{
}

\SW{
The second strategy, termed ST2, is based on the idea that the number of fine-tuning steps for each DCN should be adequately controlled so as to achieve a good trade-off between generalization and specialization abilities. This strategy is represented by 
\begin{equation}
\hat{B}^i = arg \max_{B^i} \left( 
\sum_{k=1}^i - \left| N(B_k^i) - \bar{N} \right|
+ \frac{N(B_k^i)}{1+i} 
\right).
\end{equation}
$\bar{N}$ is a pre-set integer parameter
and represents the appropriate number of fine-tunings.
In our experiments, we test three different values $\bar{N}=$1, 2, and 3.
}{
}

\SW{
The third strategy, termed ST3, is based on the idea that individual training sets are not equally important
and there must be a single most useful training set, which should be preferentially selected for the current mission's retraining. This strategy is represented by
\begin{equation}
\hat{B}^i = arg \max_{B^i} \left( 
\sum_{k=1}^i B_k^i \cdot \bar{B}^T \cdot \delta\left(N(B_k^i)-1\right)
\right).
\end{equation}
$\bar{B}$ is a pre-set vector parameter
where 
the $j$-th element is $exp(-|\bar{k}-j|)$
and $\bar{k}$ is the identifier (ID) of the appropriate training set.
In our experiments, we test all the $i$ different IDs ($\bar{k}=$1, 2, through $i$).
}{
}

\SW{
Fig. \ref{fig:C} shows different settings for the scheduling strategies described above. We considered a sequence of four seasons, three different parameter settings $\bar{N}=1,2,3$ for ST2, and four different settings $\bar{k}=1,2,3,4$ for ST3.
}{
}

\figC

\subsection{Unsupervised Place Definition}\label{sec:upd}

\SW{
The unsupervised place definition is a pre-processing part of the per-classifier fine-tuning procedure, used to partition the robot's workspace into places, so as to maximize VPC performance. A place definition algorithm takes as input a set of images and viewpoints collected by the mobile robot in the target environment. Once place classes are defined, we group images into clusters with the same place ID. Note that the place definition should occur prior to training of the classifier, and influences both training and classification performance.
}{
}

\SW{
We developed three different place definition strategies.
}{
}

\SW{
The first is location cue strategy. It partitions the sequence of images by the robot's travel distance, and assigns each sub-sequence a place label. This strategy is robust against variations in the robot's speed but does not take into account appearance information that is available from the DCN. Length of travel distance for each sub-sequence is pre-defined as a constant $T_d$. In this study, we performed a coarse optimal discretization search among $T_d=$$3i$ [m] $(i\ge 1)$, and chose $T_d=18$, which provided a good balance between efficiency and accuracy.
}{
}

\SW{
The second strategy is combined location-appearance cue strategy.
The basic idea is to use an intermediate layer's response from an independent DCN as an additional cue for clustering images into place classes.
We use the 6-th layer from a DCN $C^*$ as the visual cue, as it demonstrated excellent performance in image classification tasks in \cite{17d}. 
The workspace partitioning procedure is as follows. 
(1) Images are represented by 4,096 dimensional 6-th layer features from the DCN. 
(2) These are used as input for k-means clustering to obtain image clusters.
(3) The location cue is performed on each cluster to further partition the cluster into sub-clusters.
For the DCN $C^*$, we used 
the aforementioned $C_1^0$
that is pre-trained on
ImageNet LSVRC-2012 dataset.
}{
}

\SW{
The third strategy is an incremental clustering based on location and appearance cues. We represent appearance of a place class by a keyframe with its L2-normalized 4,096 dimensional 6-th layer feature ($f$) from the DCN, and represent location of each keyframe or each mapped image by its viewing location $(x,y)$ and viewing angle $\theta$ with respect to the global map coordinate. The clustering algorithm begins with an empty set of place classes, and then iterates for each mapped image. During each iteration, it tries to insert the mapped image into a spatially nearest place class, whose viewing location is closest to that of the mapped image. If viewing location $(x,y)$, viewing angle $\theta$ and appearance feature $f$ 
of the spatially nearest place class 
are sufficiently similar with $x_i$, $y_i$, $\theta_i$, $f_i$ of the mapped image, such that $|(x-x_i,y-y_i)|<30$, $|\theta-\theta_i|<\pi/6$ and $|f-f_i|<0.8$, it inserts the mapped image into the class. Otherwise it creates a new place class using the mapped image as the sole member. 
}{
}

\subsection{Information Fusion}\label{sec:fusion}

\SW{
The information fusion function $F$ 
takes as input a set of $|C^i|$ classifier responses
and produces a list of top-$X$ ranked place classes.
We exploit the probability value 
returned by the last layer of each DCN classifier.
The procedure begins by concatenating the 
top-$X$ ranked place classes from each DCN classifier,
to obtain a list with length $|C^i|X$.
We do not calibrate the probability distribution of individual DCN classifiers prior to the concatenation.
Then, the concatenated list is sorted in the order of highest to lowest probability value
and the top-$X$ ranked classes are output as the final classification result.
}{
}

\figG

\figE
\figF

\figBa
\figBb

\section{Experiments}

\SW{
We evaluated the suitability of the methods 
presented above
for
long-term map learning using the NCLT dataset \cite{6}. 
The NCLT dataset is a long-term autonomy dataset for robotics research collected on the University of Michigan's North Campus (Fig. \ref{fig:G}). The dataset consists of omnidirectional imagery, 3D lidar, planar lidar, GPS, and odometry data, and we use the monocular images from the front-directed camera (``camera \#5") for our VPC tasks. During vehicle travel through both indoor and outdoor environments, various types of appearance changes are encountered with respect to the mapped images. These originate from the movement of people, parked cars, furniture, construction of the building, opening/closing of doors, placing/removing of posters, as well as other nuisance changes originating from illumination changes, viewpoint dependent changes of object appearances and occlusions, weather changes, falling leaves and snow. These appearance changes make our cross-season VPC task a challenging one. 
We repeated the long-term map learning in Fig. \ref{fig:A} four times (See Fig. \ref{fig:C}), by using four datasets from four different seasons ``2012/3/31," ``2012/8/04," ``2012/11/17," and ``2012/1/22"  as individual training sets, and an additional set ``2012/2/19" as test data for the last (i.e., 4-th) mission. 
}{
}
\SW{
We followed a standard procedure for fine-tuning. The classification function in the DCN is a softmax classifier that computes the probability of all the place classes. To fine-tune the DCN, we changed the softmax classifier using a new value equal to the number of place classes. The DCN parameters were then fine-tuned on the new training datasets. 
Input images were resized to 256 $\times$ 256.
The DCN parameters were then fine-tuned on the new training datasets. 
Fig. \ref{fig:G} shows a bird's eye view of the environment and the robot's trajectories of the four adaptation missions. 
}{
}

\SW{
Fig. \ref{fig:E} shows performance results. We conducted performance evaluations for the different UPD algorithms described in \ref{sec:upd}: 
location cue strategy (``\#1"),
location-appearance cue strategy (``\#2"), and
incremental clustering strategy (``\#3").
We also conducted performance evaluations for two different VPC scenarios, ``fine localization" and ``coarse localization", in which allowed localization errors were set to 10 m and 20 m, respectively. We also considered a different type of test data, which is identified by ``test:ex". Unlike the default setting where the $(i+1)$-th exploration season's set is used as test data for the $i$-th adaptation mission, the setting ``test:ex" uses a fixed test set ``2012/2/19" regardless of the mission ID ($i$). 
}{
}
\SW{
Note that
the scheduling strategy ST2
with $\bar{N}=1$ 
is competitive or outperforms the other strategies
for almost all missions
and for both the fine and coarse localization scenarios
as well as for both types of test data.
As mentioned,
this strategy controls 
fine-tuning number
as close to $\bar{N}$ as possible
so as to achieve a good trade-off between generalization and specialization abilities. 
Moreover,
the appropriate parameter $\bar{N}$ 
turned out to be 1,
meaning that 
in the case of ST2,
fine-tuning should be performed only once for each DCN.
The reason may be that 
fine-tuning more than once 
led to over-fitting
and could not generalize well to the unseen test data.
Among the other strategies,
ST3 with $\bar{k}=1$ exhibited relatively good performance.
The reason may be that 
the single DCN trained on the specific season $\bar{k}=1$ (``2012/3/31") 
was well-suited for much of the test data considered here.
From the above results,
it could be concluded that 
the proposed framework of 
planned retraining scheduling
combined with information fusion
is effective for cross-season VPC tasks,
particularly when
fine-tuning number is controlled.
}{
}

\SW{
Figs. \ref{fig:Ba} and \ref{fig:Bb} show success and failure examples. 
We used strategy ST2 with $\bar{N}=1$ for the ensemble classifier.
As shown in Fig. \ref{fig:Ba},
the classifier captures
scene structure and 
discriminative characteristics of the scenes 
both for indoor and outdoor environments.
On the other hand,
failure often occurs from non-discriminative scenes
as shown in Fig. \ref{fig:Bb}.
}{
}

\SW{
Fig. \ref{fig:H} shows instances of unsupervised place definition.
We show results for three different definition algorithms.
As can be seen,
the location cue strategy 
uniformly partitioned the robot's trajectories into equal-length sub-trajectories (i.e., place classes).
On the other hand, 
the location-appearance cue strategy and the incremental clustering strategy tend to group similar successive locations into the same class.
These two strategies yielded the best performances 
and the former was slightly better than the latter
in the experiments conducted (See Figs. \ref{fig:E} and \ref{fig:F}).
}{
}

\figH

\section{Conclusions}

\SW{
We presented a long-term map learning framework for cross-season VPC.
This framework enabled efficient transfer learning from one season to the next, at a small constant cost, and without wasting the robot's available long-term-memory by memorizing very large amounts of training data. To realize an acceptable tradeoff between generalization and specialization abilities, we employed an ensemble of DCN classifiers and considered the task of scheduling when and which classifiers to retrain, given a previous season's DCN classifiers as the sole prior knowledge. We also presented a unified framework and proposed practical strategies to implement retraining scheduling. Furthermore, we addressed the task of partitioning the robot's workspace into places to define place classes in an unsupervised manner, to maximize VPC performance. Through long-term map learning and VPC experiments, we have shown that (a) the ensemble DCN classifier performs comparably or better than a single DCN classifier, and (b) retraining scheduling of DCN classifiers is crucial, to achieve a good balance between generalization and specialization.
}{
}

\SW{
Future work should address the map building stage. Currently, our experimental implementation assumes fine-grained viewpoint information for mapped images and future work should focus on the issue of map errors. Furthermore, visual place classifiers should be modified when viewpoint information of mapped images is incrementally updated during the long-term multi-session map building process. Adaptation of the place definition to changing environments is another important direction for future research.
}{
}

\bibliographystyle{IEEEtran}
\bibliography{upd}

\end{document}